\documentclass{SVProc}
\usepackage{makeidx}  % allows for indexgeneration
\makeindex
\usepackage{url}

\usepackage[english]{babel}
\usepackage[utf8]{inputenc}
\usepackage[T1]{fontenc}
\usepackage{amssymb}
\usepackage{amsmath}
\usepackage{amsfonts}
\usepackage[pdftex]{graphicx}
\usepackage{todonotes}
\usepackage{hyperref}

\newcommand{\R}{\mathbb R}

\newcommand{\DQ}{\mathbb{DH}}

\newcommand{\qi}{\mathbf{i}}
\newcommand{\qj}{\mathbf{j}}
\newcommand{\qk}{\mathbf{k}}
\newcommand{\e}{\varepsilon}
\newcommand{\Cj}[1]{{#1}^\ast}

\newcommand{\norm}[1]{\left\|#1\right\|}

\begin{document}

% \frontmatter          % for the preliminaries
% %
% \pagestyle{headings} 
% switches on printing of running heads
%%%\addtocmark{Hamiltonian Mechanics} % additional mark in the TOC
%
% \chapter*{Preface}
% %
% This textbook is intended for use by students of physics, physical
% chemistry, and theoretical chemistry. The reader is presumed to have a
% basic knowledge of atomic and quantum physics at the level provided, for
% example, by the first few chapters in our book {\it The Physics of Atoms
% and Quanta}. The student of physics will find here material which should
% be included in the basic education of every physicist. This book should
% furthermore allow students to acquire an appreciation of the breadth and
% variety within the field of molecular physics and its future as a
% fascinating area of research.

%

%
%\section*{Sponsoring Institutions}
%
% V. Meyer Inc., Reading, MA, USA\\
% The Hofmann-International Company, San Louis Obispo, CA, USA\\
% Kramer Industries, Heidelberg, Germany
%
%\tableofcontents
%
\mainmatter              % start of the contributions
%
%\part{Hamiltonian Mechanics}
%
\title{An Overconstrained Vertical Darboux Mechanism}
\author{Johannes Siegele \and Matrin Pfurner}

\titlerunning{An Overconstrained Vertical Darboux Mechanism}  % abbreviated title (for running head)
\authorrunning{Johannes Siegele and Martin Pfurner} % abbreviated author list (for running head)
\tocauthor{Johannes Siegele, Martin Pfurner}
\index{Pfurner, M.}
\index{Siegele, J.}
\institute{Department of Basic Sciences in Engineering, Universit\"at Innsbruck, Innsbruck, Austria\\
  \email{Johannes.Siegele@uibk.ac.at, Martin.Pfurner@uibk.ac.at}
}

\maketitle              % typeset the title of the contribution

\begin{abstract}
  In this article, we will construct an overconstrained closed-loop linkage
  consisting of four revolute and one cylindrical joint. It is obtained by
  factorization of a prescribed vertical Darboux motion. We will investigate the
  kinematic behaviour of the obtained mechanism, which turns out to have
  multiple operation modes. Under certain conditions on the design parameters,
  two of the operation modes will correspond to vertical Darboux motions. It
  turns out, that for these design parameters, there also exists a second
  assembly mode.

\end{abstract}

\section{Introduction}

% The vertical Darboux motion is a special type of motions classified by Darboux
% in 1881 with the property, that every point has a planar trajectory. Such a
% motion is called vertical, if it is the composition of a rotation with a
% suitably parametrized oscillating translation in the direction of the rotation axis.

In 1881, Darboux determined all possible motions with the property, that every
point has a planar trajectory. Vertical Darboux motions are a sub-type of these
motions and are obtained by the composition of a rotation with a suitably
parametrized oscillating translation in the direction of the rotation axis. All
generic point trajectories for both the non-vertical and the vertical Darboux
motion are ellipses. The vertical Darboux motion is in addition a cylindrical
and line symmetric motion. For more detail we refer to \cite[Chapter
9]{bottema90}.

Vertical Darboux motions are of particular interest, when using Study parameters
for the representation of spatial displacements. Any line in the ambient space
of the Study quadric represents a vertical Darboux motion \cite{purwar10}. Lines on
the Study quadric correspond to rotations and translations. Therefore the
vertical Darboux motion is a natural generalization of rotations and
translations. By representing the motion by a curve on the Study quadric,its
instantaneous behaviour corresponds to the instantaneous motion given by the
curve tangent, which is a line in the ambient space. Thus, vertical Darboux
motions may also be used for the description of the instantaneous behaviour of a
motion.

In this article, we construct an overconstrained $4$RC-linkage performing an
arbitrary vertical Darboux motion. The construction is based on the
factorization theory for dual quaternion polynomials \cite{hegedus13} and on the
construction of a non-vertical Darboux linkage in \cite{li15}. Lines on the
Study quadric can be parametrized by linear dual quaternion polynomials, thus
they represent rotations or translations. Both motions can easily be realized by
revolute or prismatic joints, respectively. Thus, by decomposing a dual
quaternion polynomial into the product of linear factors, we are able to
construct open kinematic chains.
For the vertical Darboux motion, we obtain an open chain, which can perform a
cylindrical motion. Therefore, it can be closed using a cylindrical joint to
obtain a single-loop mechanism.

Overconstrained mechanisms performing a vertcal Darboux motion are constructed
in \cite{Lee12,Lee15}. Our approach, however, yields a new type of
overconstrained mechanisms. We will analyze operation and assembly modes of the
obtained linkage. In general, they will have two operation modes, one of them is
the desired vertical Darboux motion, the other is a cylindrical motion of degree
5. Further, we will give a condition, which ensures existence of a second
assembly mode, as well as the decompositon of the second operation mode into
another vertical Darboux motion and two rotations.

% A
% general Darboux motion is parametrized by a cubic dual quaternion polynomial
% which can be decomposed into the product of three linear factors which
% corresponds to an open $3$R-chain and F

\section{Preliminaries}

In this manuscript, we will construct a closed-loop linkage able to perform a
vertical Darboux motion. Our construction is based on the factorizaton theory of
dual quaternion polynomials, therefore we will give a short introduction to dual
quaternions and motion polynomials in this section. For further detail we refer
to \cite{Li2015}.

\subsection{Dual Quaternions}
A dual quaternion $h\in\DQ$ is given by
\begin{equation*}
  h=p_0+p_1\qi+p_2\qj+p_3\qk+d_0\e+d_1\e\qi+d_2\e\qj+d_3\e\qk
\end{equation*}
for real numbers $p_0,\ldots,p_3$, $d_0,\ldots,d_3\in\R$. The non-commutative
multiplication of dual quaternions abides by the rules
\begin{equation*}
  \qi^2=\qj^2=\qk^2=\qi\qj\qk=-1,\quad \e^2=0,\quad \e\qi=\qi\e,\quad\e\qj=\qj\e,\quad\e\qk=\qk\e.
\end{equation*}
The quaternions $p=p_0+p_1\qi+p_2\qj+p_3\qk$, $d=d_0+d_1\qi+d_2\qj+d_3\qk$ are
called primal and dual part of $h$. The dual quaternion conjugate is given by 
\begin{equation*}
 \Cj{h}=p_0-p_1\qi-p_2\qj-p_3\qk+d_0\e-d_1\e\qi-d_2\e\qj-d_3\e\qk,
\end{equation*}
the dual quaternion norm is given by $\norm{h}=h\Cj{h}$. Dual quaternions can be
used to represent rigid body displacements by simply using the Study parameters
of a displacement as the coefficients of the dual quaternion. The action of a
displacement on a point $[x_0,x_1,x_2,x_3]$ in projective three-space can be
represented by a dual quaternion product by embedding the point into the dual
quaternions via $[x_0,x_1,x_2,x_3]\mapsto x_0+\e x$ with
$x=x_1\qi+x_2\qj+x_3\qk$. Acting on this point by a displacement given by $p+\e
d$ corresponds to computing the product
\begin{equation*}
  (p-\e d)(x_0+\e x)(\Cj{p}+\e\Cj{d}).
\end{equation*}
The coefficients of a dual quaternion $h$ fulfill the Study condition if and
only if $\norm{h}$ is real. Note that all scalar multiples of a dual quaternion
yield the same displacement.

\subsection{Motion Polynomials}

Rational motions can be represented by polynomials with dual quaternion
coefficients $Q=\prod_{\ell=0}^n q_\ell t^\ell$ with $q_0$, $q_1,\ldots,q_n\in\DQ$ such
that $\norm{Q}=Q\Cj{Q}\in\R[t]$ is a real polynomial. Here the conjugate
polynomial $\Cj{Q}$ is obtained by conjugating all of its coefficients. Such
polynomials are called motion polynomials.

The simplest examples of motion polynomials are linear, monic polynomials $t-h$,
where the scalar coefficient $d_0$ of the dual part has to vanish for the Study
condition to be fulfilled. Such a linear polynomial either represents a
rotation, if $\norm{t-h}$ has complex roots, or a translation otherwise. In case
of a rotation, its axis has Pl\"ucker coordinates
$[p_1,p_2,p_3,-d_1,-d_2,-d_3]$. Otherwise the direction of translation is given
by $[d_1,d_2,d_3]\in\R^3$. Both of these motions can be realized by revolute or
prismatic joints, respectively. Decomposing a given motion polynomial into the
product of linear factors therefore corresponds to decomposing the represented
motion into a concatenation of rotations and translations, which in turn can be
realized by joints. This gives rise to a kinematic chain which is able to
perform the given motion. It can be constrained by another chain generated by a
different factorization of the same motion polynomial, which yields a closed
mechanism still able to perform the given motion.

\section{Vertical Darboux Motion}\label{sec2}

%%%%%%%%%%%%%%%%%%%%%%%%%%%%%%%%%%%%%%%%%%%%%%%%%%%%%%%%%%%%%%%%%%%%%%%%%%%%% 
A vertical Darboux motion is the composition of a rotation and an oscillating
translation along the same axis. Vertical Darboux motions around the third
coordinate axis can be parameterized by the dual quaternion polynomial
\cite{Li2015}.
\begin{equation*}
M = (t^2+1)(t-\qk)+\e(-b\qk t + c\qk)(t-\qk).
\end{equation*}
It does not admit a factorization into three linear factors, but multiplying $M$
with~$(t^2+1)$ allows us to find factorizations, each consisting of 5 linear
polynomials \cite{Li19}. Every factorization corresponds to an open kinematic
chain with at most five revolute joints, which can perform the vertical Darboux
motion given by $M$. Combining several of these chains would result in a rather
complicated mechanism. But since the vertical Darboux motion is a cylindrical
motion, we can close the obtained open chain with a C-joint. This results in a
closed-loop mechanism with at most six joints where one of the joints has two
degrees of freedom. To obtain an overconstrained mechanism, we can try to find
factorizations for which two neighboring factors are equal. This yields an open
$4$R chain which can be close with a C-joint resulting in an overconstrained
mechanism.

\subsection{Factorization of the vertical Darboux motion}
Like in \cite[Section 3.3]{li15}, we will try to find $P_4\in\DQ[t]$ such that
$P_4^2$ is a right factor of $(t^2+1)M$. Solving a system of equations for the
coefficients of $P_4$ shows that it has to be of the shape
\begin{equation*}
P_4=t-k-\e(q_1\qi+q_2\qj)
\end{equation*}
for arbitrary $q_1$, $q_2\in\R$. After dividing off these two
right factors, we obtain
\begin{equation*}
Q=(t^2+1)(t+\qk)+\e(bt-c+2t(q_1t+q_2)\qi-2t(q_2t+q_1)\qj-t(bt-c)\qk),
\end{equation*}
which represents a Darboux motion. As long as $q_1$ and $q_2$ do not vanish
simultaneously, it is non-vertical, thus admits infinitely many factorizations
into three linear factors. Using factorization techniques, it is straight forward to compute
\begin{equation*}
P_3=t+\qk+\e(y_1\qi+y_2\qj),
\end{equation*}
which is a right factor of $Q$. Dividing off this factor will leave us with a
quadratic translation, which admits a factorization if and only if it is a
circular translation~\cite{Li2015}. To ensure factorizability, we need to choose
\begin{align*}
  y_1&=\frac {b^2 q_1-2bc q_2-c^2 q_1+4 q_1^3+4q_1q_2^2}{4(q_1^2+ q_2^2)},\\
  y_2&=\frac {b^2q_2+2bcq_1-c^2q_2+4 q_1^2q_2+4q_2^3}{4(q_1^2+q_2^2)}.
\end{align*}
The resulting motion is then a translation along a circle with axis in the
direction $[4(bq_1-cq_2),4(bq_2+cq_1),-b^2-c^2+4q_1^2+4q_2^2]$. To find a
factorization for this translation, we can simply take any line parallel to the
circle axis and use its normalized Pl\"ucker coordinates as the coefficients of
the right factor, i.e. we can define
\begin{equation*}
  P_2=t+\frac{4(bq_1-cq_2)\qi+4(bq_2+cq_1)\qj-(b^2+c^2-4q_1^2-4q_2^2)\qk}{b^2+c^2+4q_1^2+4q_2^2}-\e(z_1\qi+z_2\qj+z_3\qk)
\end{equation*}
for $z_1$, $z_2$, $z_3\in\R$ such that the Study (Pl\"ucker) condition is
fulfilled. After dividing off $P_2$ we are left with the last factor which is
given by
\begin{align*}
  P_1=t&-\frac{4(bq_1-cq_2)\qi+4(bq_2+cq_1)\qj-(b^2+c^2-4q_1^2-4q_2^2)\qk}{b^2+c^2+4q_1^2+4q_2^2}\\
  &+\e \frac{
    2 q_2 \left( bc+2 q_1 q_2+2 z_1 q_2\right) - q_1 \left( b^2-c^2-4 q_1^2-4 q_1 z_1 \right)}{4q_1^2+4q_2^2}\qi\\
   &-\e\frac{2 q_1 \left( bc-2 q_1 q_2-2 q_1 z_2\right) + q_2\left( b^2-c^2-4
       q_2^2-4  q_2 z_2 \right)}{4q_1^2+4q_2^2}\qj\\
   &+\e\frac{4(bq_1-cq_2)z_1+4(bq_2+cq_1)z_2-b(b^2+c^2-4q_1^2-4q_2^2)}{b^2+c^2-4q_1^2-4q_2^2}\qk
\end{align*}
assuming ${b^2+c^2-4q_1^2-4q_2^2}\neq 0$. Note, that $z_3$ is chosen such that
the Study condition for $P_2$ is fulfilled.

\begin{figure}[b!]
  \centering
  \includegraphics[width=0.7\linewidth]{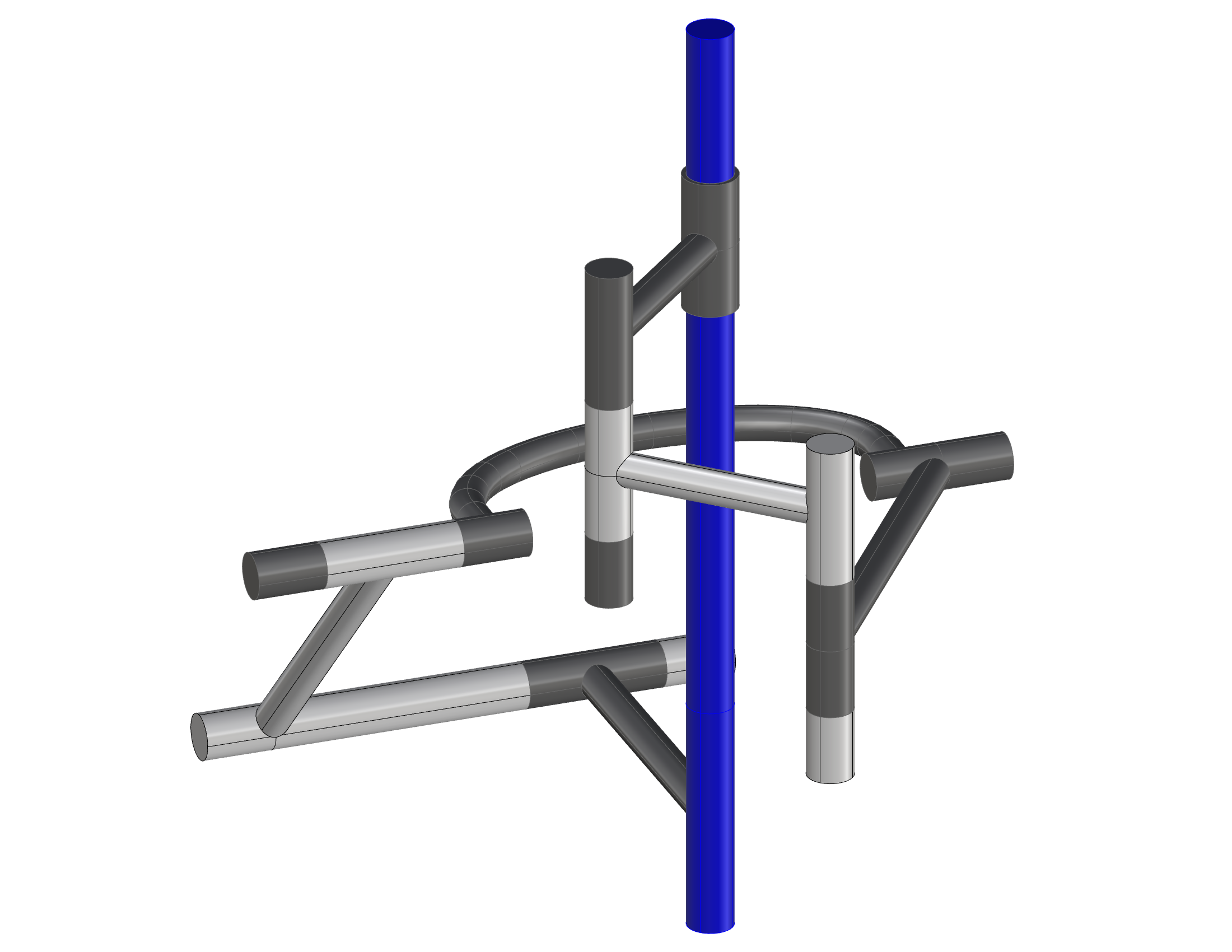}
  \caption{An example of a 4RC vertical Darboux mechanism (C-joint is blue).}
  \label{fig:0}
\end{figure}

If ${b^2+c^2-4q_1^2-4q_2^2}= 0$,
$z_3$ can be chosen arbitrarily, with the restriction that $z_1$ and $z_2$ need
to fulfill $[z_1,z_2]=z[bq_2+cq_1,-bq_1+cq_2]$ for arbitrary $z\in\R$. With this
condition, the last factor is given by
\begin{align*}
  P_1=t&-2\frac{(bq_1-cq_2)\qi+(bq_2+cq_1)\qj}{b^2+c^2}\\
      &-\e\frac{(b^2+c^2)z-2c}{b^2+c^2}((bq_2+cq_1)\qi-(bq_1-cq_2)\qj)\\
  &+\e(z_3-b)\qk.
\end{align*}
This factorization
\begin{equation*}
  (t^2+1)M=P_1P_2P_3P_4^2
\end{equation*}
now gives rise to a $4$R chain, which we can close with a cylindrical joint to
obtain a closed-loop linkage, see Fig,~\ref{fig:0}. It admits, by construction,
the initial vertical Darboux motion given by $M$ as one operation mode.

\section{Kinematic Analysis of the Vertical Darboux Mechanism}
To analyze other possible operation modes, let us investigate the kinematic
chain obtained by these factors, where each joint can move inepenently of each
other, i.e. $C =P_1(v_1)P_2(v_2)P_3(v_3)P_4(v_4)(\tau-k)(1-\e s\qk)$. Here the
last two factors simply describe a cylindrical joint with the third coordinate
axis as joint axis. A kinematic chain can be closed, if the third coordinate
axes of the base and the moving frame coincide. This yields two closure
conditions, the first one being, that the axes point in the same direction, the
second one, that they point in opposite directions.

\subsection{First Assembly Mode}

The first closure condition means that the coefficients of the dual quaternion
units $\qi$, $\qj$, $\qk$, $\e$, $\e\qi$, $\e\qj$ and $\e\qk$ of $C$ vanish,
i.e. $C$ describes the identity transformation. This gives us seven polynomial
equations, where the first and the second have a common factor $v_1-v_2$ while
the other factors do not have common real solutions. After substituting this
into our set of equations, the third equation has the factor $\tau v_3-\tau
v_4+v_3v_4+1$ while last equation has the factor $sv_1^2+bv_1-c+s$. They admit
the real solution $s=-(bv_2-c)/(v_1^2+1)$, $\tau=-(v_3v_4+1)/(v_3-v_4)$. After
substituting these solutions into the remaining equations, we are left with two
polynomial equations which are quadratic in each of the variables $v_1$, $v_3$
and $v_4$. Computing a resultant to eliminate $v_4$ and dividing off unnecessary
factors yields an equation with two factors, one of them is $v_1-v_3$, the other
\begin{align}
  F =&8bc q_1^2 v_1^3 v_3+8bc q_2^2 v_1 ^3 v_3+b^4 v_1^3-b^4 v_1^2 v_3+2 b^2c^2
       v_1^3-2b^2c^2 v_1^2 v_3+4 b^2 q_1^2 v_1^3\notag\\
     &+12b^2 q_1^2 v_1^2 v_3+4b^2 q_2^2v_1^3 +12b^2 q_2^2 v_1^2 v_3+c^4 v_1^3-c^4 v_1^2 v_3-4c^2 q_1^2 v_1^3\notag\\
     &-12c^2 q_1^2 v_1^2 v_3-4c^2 q_2^2 v_1^ 3-12c^2 q_2^2 v_1^2 v_3-24bc q_1^ 2 v_1^2-24bc q_1^2 v_1
       v_3\notag\\
     &-24bc q_2 ^2 v_1^2-24bc q_2^2 v_1 v_3+ v_1 b^4-b^4 v_3+2 v_1c^2b^2-2b^2c^2
       v_3-12b^2 q_1^2 v_1\nonumber\\
     &-4b^2 q_1^2 v_3-12b^2 q_2^2 v_1-4b^2 q_2^2 v_3+ v_1c^4-c^4
       v_3 +12c^2 q_1^2 v_1+4c^2 q_1^2 v_3\nonumber\\
     &+12c^2 q_2^2 v_1+4c^2 q_2^2 v_3+8bc q_1^2+8bc
       q_2^2\label{eq:F}
       \end{align}
This, in general, gives rise to two sets of solutions, the first one being
$v_3=v_1$ which in turn also yields $v_4=(v_1^2-1)/2v_1$. This solution
corresponds to the initial vertical Darboux motion.

\begin{figure}
  \centering
  \includegraphics[]{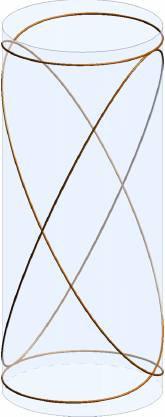}\qquad
  \includegraphics[]{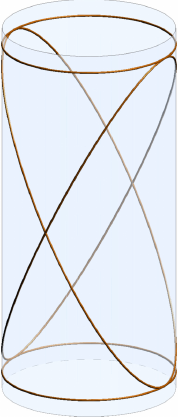} 
  \caption{Trajectories of the coupler motion for $b=1$, $c=2$ and $b=c=\sqrt{2}$. }
  \label{fig:1}
\end{figure}

The second solution is obtained by solving $F$ for $v_3$ as it is linear in this
variable and resubstituting the obtained solution into the system of equations.
This yields two equations with a common factor linear in $v_4$ and each of them
has one other factor, respectively, which do not have a common solution provided
$b^2+c^2-4q_1^2-4q_2^2\neq 0$ (this case will be investigated below). This
common factor yields the solutions
\begin{align*}
  v_3&=\frac{v_1(v_1^2+1)(b^2+c^2)^2+\left((b^2-c^2)(v_1^3-3v_1)-6bcv_1^2+2bc\right)(4q_1^2+4q_2^2)}{
       (v_1^2+1)(b^2+c^2)^2 -((b^2-c^2)(3v_1^2-1)+2bcv_1^3-6bcv_1)(4q_1^2+4q_2^2)     
       }
  \\
  v_4&=-\frac {( b v_1+c v_1+b-c)( bv_1-c v_1-b-c)}{2(c v_1+b )( b v_1-c )}.
\end{align*}
This solution corresponds to a motion with trajectories of degree six. It
is the composition of the vertical Darboux motion given by
\begin{equation*}
  (t^2+1)(bt-c-(ct+b)\qk)+\e(bt-c)(ct+b+(bt-c)\qk)
\end{equation*}
and a quadratically parametrized rotation around the third coordinate axis
\begin{equation}\label{eq:vdarb2}
  -(bt-2ct-b)(b^2+c^2+4q_1^2+4q_2^2) + (ct^2+2bt-c)(b^2+c^2-4q_1^2-4q_2^2)\qk.
\end{equation}
Figure~\ref{fig:1} (left) shows a trajectory of the vertical Darboux motion
(first solution) and this other motion (second solution) for the values $q_1=1$,
$q_2=0$, $z_1=0$, $z_2=0$.

Let us now investigate the case, where $b^2+c^2-4q_1^2-4q_2^2= 0$. With this
condition the factor $F$ in Eq.~\eqref{eq:F} of the resultant simplifies to
\begin{align*}
  F=(bv_1^2-2cv_1-b)(cv_1v_3+bv_1+bv_3-c).
\end{align*}
The second factor yields the same solutions as in the case above. The first
factor, however, gives rise to two additional sets of solutions
\begin{equation*}
  v_1=\frac{c\pm\sqrt{b^2+c^2}}{b},\qquad  v_4=\frac{c}{b}.
\end{equation*}
Both of these solutions correspond to rotations around the third coordinate axis
given by
\begin{equation*}
  2(c\sqrt{b^2+c^2}\pm(b^2+c^2))(ct+b-(bt-c)\qk)-\e b^2\sqrt{b^2+c^2}(bt-c+(ct+b)\qk).
\end{equation*}
Further, the polynomial in Eq.~\eqref{eq:vdarb2} simplifies to a real
polynomial, which implies, that the second solution in this special case also
corresponds to a vertical Darboux motion. The trajectories of all of these
motions are depicted in Figure~\ref{fig:1} (right).

\subsection{Second Assembly Mode}

For the second closure condition, the dual quaternion coefficients of $1$,
$\qk$, $\e$, $\e\qi$, $\e\qj$, $\e\qk$ need to vanish, while the coefficients of
$\qi$ and $\qj$ must fulfill a linear equation. This corresponds to assembling
the open kinematic chain such that the third coordinate axes of the base and
moving frame coincide, but they point in opposite directions. For the linear
condition of the coefficients $c_i$ and $c_j$ of $\qi$ and $\qj$ we will use the
equation $(bq_1-cq_2)c_i+(bq_2+cq_1)c_j=0$ which will be the second equation in
our closure condition.

Solving the first equation for $v_2$ and substituting the result into the third
equation yields an equation which simplifies, after dividing off unnecessary
factors, to $b^2+c^2-4q_1^2-4q_2^2=0$. Thus, this second assembly mode only
exists, if this condition is fulfilled. Note that this condition is the same as
in the section above for the existence of four operation modes in the first
assembly.

In this case, the first, second and third equation only have one common
solution for $v_2$ and $\tau$, namely
\begin{equation*}
v_2=-1/v_1,\qquad \tau=(v_3-v_4)/(v_3v_4+1).
\end{equation*}
After resubstituting
these solutions, equations five and six have one common factor which is linear
in $s$ while their other factors do not admit common solutions. The solution for
$s$ is
\begin{equation*}
  s=\frac{z(b^2+c^2)(v_1^2+1) + 2bv_1 - 2c}{2(v_1^2+1)}.
\end{equation*}
After resubstituting this solution, the last equation, after dividing off
unnecessary factors, reads
\begin{equation}\label{eq:2}
  (bv_4-c)(v_3^2 - 2v_3v_4 - 1)=0.
\end{equation}
The first factor in Eq.~\eqref{eq:2} yields the two solutions
\begin{equation*}
  v_1 = \frac{-c\pm\sqrt{-3b^2+8bz_3+c^2-4z_3^2}}{b-2z_3},\qquad v_4=\frac{c}{b}.
\end{equation*}

For the second factor in Eq.~\eqref{eq:2} we get $v_4=(v_3^2-1)/2v_3$.
Resubstituting this solution yields the equation
\begin{equation*}
   -v_1^2v_3^2z_3 + cv_1^2v_3 + cv_1v_3^2 + bv_1^2 + bv_3^2 - v_1^2z_3 - v_3^2z_3 + cv_1 + cv_3 + 2b - z_3=0
\end{equation*}
This equation is quadratic in $v_1$ (and $v_3$), thus solving it for $v_1$ will
yield two solutions.
% Thus we obtain two solutions given by
% \begin{align*}
%   v_1=-\frac{
%   cv_3^2 + c \pm \sqrt{4bv_3^4z_3 + c^2v_3^4 - 4v_3^4z_3^2 - 4bcv_3^3 + 8cv_3^3z_3 - 4b^2v_3^2 + 16bv_3^2z_3 - 2c^2v_3^2 - 8v_3^2z_3^2 - 12bcv_3 + 8cv_3z_3 - 8b^2 + 12bz_3 + c^2 - 4z_3^2}
%   }{2(-v_3^2z_3+cv_3+b-z_3)}
% \end{align*}

\begin{figure}[t!]
  \centering
  \includegraphics[]{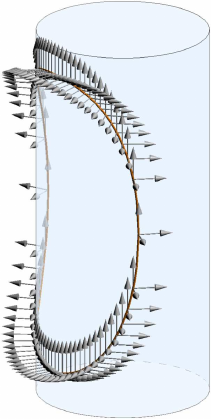}\qquad
  \includegraphics[]{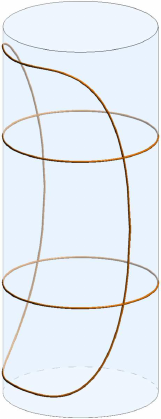} 

  \caption{Coupler motion where the first two solutions are
    complex (left) or real (right).}
  \label{fig:2}
\end{figure}

In contrast to the first assembly mode, all solutions depend on $z$ and $z_3$,
but not on $q_1$ and $q_2$. Further they contain square roots, thus the
solutions can be complex. On the left hand side of Fig.~\ref{fig:2} the
trajectories of a point under these motions are shown for $b=c=\sqrt{2}$,
$z=z_3=0$. For these values, only the second operation mode admits real
trajectories. On the right hand side of Fig.~\ref{fig:2}, the trajecories for
$b=1$, $c=2$, $z=z_3=0$ are shown. Here, also the first two solutions are real
and the corresponding motions are rotations around the third coordinate axis.

\section{Conclusion}
We have generated an overconstrained 4RC closed-loop linkage, which is able to
perform a prescribed vertical Darboux motion. Its kinematic analysis revealed
the existence of, in general, two operation modes, one of them corresponding to
the initial vertical Darboux motion. We gave a condition on the design
parameters of the mechanism for which the second operation mode decomposes into
two rotations and an additional vertical Darboux motion. The same condition also
ensures the existence of a second assembly mode, which in turn has up to three
real operation modes.

\section*{Acknowledgement} 

Johannes Siegele was supported by the Austrian Science Fund (FWF): P~33397 (Rotor
Polynomials: Algebra and Geometry of Conformal Motions).


\bibliographystyle{plain}
\bibliography{Bib}

\clearpage
%\addtocmark[2]{Author Index} % additional numbered TOC entry
\renewcommand{\indexname}{Author Index}
%\printautindex

\end{document}